\documentclass{article}
\usepackage{ijcai21}

\usepackage{times}
\usepackage{xcolor}

\usepackage{soul}
\usepackage{url}
\usepackage[hidelinks]{hyperref}
\usepackage[utf8]{inputenc}
\usepackage[small]{caption}
\usepackage{graphicx}
\usepackage{amsmath,amsfonts,amsthm,bm}
\usepackage{booktabs}
\usepackage{algorithm}
\newtheorem{definition}{Definition}

\title{A Taxonomy of Similarity Metrics for Markov Decision Processes}

\author{
Álvaro Visús$^1$\and
Javier García$^1$\and
Fernando Fernández$^1$\\
\affiliations
$^1$Departamento de Inform\'atica, Universidad Carlos III de Madrid\\
 Avda. de la Universidad, 30. 28911 Legan\'es (Madrid). Spain
\emails
avisus@pa.uc3m.es, \{fjgpolo, ffernand\}@inf.uc3m.es
}

\begin{document}

\maketitle

\begin{abstract}
Although the notion of task similarity is potentially interesting in a wide range of areas such as curriculum learning or automated planning, it has mostly been tied to transfer learning. Transfer is based on the idea of reusing the knowledge acquired in the learning of a set of source tasks to a new learning process in a target task, assuming that the target and source tasks are \textit{close enough}. In recent years, transfer learning has succeeded in making Reinforcement Learning (RL) algorithms more efficient (e.g., by reducing the number of samples needed to achieve the (near-)optimal performance). Transfer in RL is based on the core concept of \textit{similarity}: whenever the tasks are \textit{similar}, the transferred knowledge can be reused to solve the target task and significantly improve the learning performance. Therefore, the selection of good metrics to measure these similarities is a critical aspect when building transfer RL algorithms, especially when this knowledge is transferred from simulation to the real world. In the literature, there are many metrics to measure the similarity between MDPs, hence, many definitions of \textit{similarity} or its complement \textit{distance} has been considered. In this paper, we propose a categorization of these metrics and analyze the definitions of \textit{similarity} proposed so far, taking into account such categorization. We also follow this taxonomy to survey the existing literature, as well as suggesting future directions for the construction of new metrics.
\end{abstract}

\section{Introduction}

Markov decision processes (MDPs) are a common way of encoding decision making problems in Reinforcement Learning (RL) tasks~\cite{sutton2011}. 
In RL, an MDP is considered to be solved when a policy (i.e., a way of behaving for each state) has been discovered which maximizes a long-term expected return. However, although RL is known as an effective machine learning technique, it might perform poorly in complex problems, leading to a slow rate of convergence. This issue magnifies when facing realistic continuous problems where the curse of dimensionality is inevitable. Transfer learning in RL is a successful technique to remedy such a problem. Speciﬁcally, rather than learning a new policy for every MDP, a policy could be learned on one MDP, then transferred to another, similar MDP, and either used as is, or treated as a starting point from which to learn the new policy. Clearly this transfer cannot be done successfully between any two MDPs, but only in the case they are \textit{similar}.

Therefore, in this context one question arises: when are two MDPs \textit{similar}? We must give a better definition of what constitutes \textit{similar} MDPs. In this paper, we consider the concept of \textit{similar} is related to the notion of ``positive transfer''~\cite{taylor2009transfer}. Formally, positive transfer happens when the knowledge in the source task contribute to the improved performance of learning in the target task, and it is considered a negative transfer otherwise, i.e., when the transfer hurts the learning performance when compared with learning from scratch. Additionally, the greater the improvement in the target task, i.e., the greater the positive transfer, the more similar the tasks has to be considered. It is important to be aware of the fact that, based on this description, the concept of \textit{similarity} might not be related with the \textit{structural} similarities between the MDPs. So the correct selection of metrics that allow us to measure the similarity between MDPs is a critical issue in transfer learning, precisely to avoid the negative transfer. 


The literature in transfer learning has proposed different metrics to measure the level of similarity between MDPs, hence, different definitions of the concept of \textit{similarity} have been considered so far. This paper surveys the existing similarity metrics and contributes a taxonomy that, in its root, classifies them into two clearly distinct categories: \textit{model-based}, and \textit{performance-based} metrics. We consider such a distinction as a core contribution, allowing to categorize metrics in a novel and useful way. Model-based metrics are based on the \textit{structural} similarities between the MDP models. Such model-based metrics can be computed in different ways depending on what elements of the MDP models come into play to compute the similarity~\cite{ammar14,taylor2008autonomous,milner1980,castro2011automatic,svetlik17}. 
Instead, performance-based metrics are computed by comparing the \textit{performance} of the learning agents in the source task and the target task. Such a performance comparison can be done in two different ways: by comparing the resulting policies from learning in the source task and the target task~\cite{1555955,Karimpanal2018} or, from a transfer point of view, by measuring the reuse gain, i.e., the positive transfer~\cite{mahmud13,sinapov15,fernandez12}. Many metrics can be used to measure such a reuse gain of transfer (e.g., jumpstart, asymptotic performance,  total reward)~\cite{taylor2009transfer}. In some ways, this reuse gain could be the best method for measuring similarity between two tasks~\cite{1555955}. Unfortunately, it is often difficult to compute all of these performance-based measures before actually solving the target task, since most of them require to be computed \textit{a posteriori}, i.e., after the learning processes. However, there are some few exceptions to this rule, within which, for example, the similarity is computed on-line, i.e., \textit{during} solving the target task~\cite{fernandez12}. 


We hope our taxonomy applies to a wide range of researchers, not just those interested in transfer learning. For example, the proposed metrics could also be a critical step forward to sort the samples and tasks in \textit{Curriculum Learning}~\cite{narvekar2020curriculum}, or could be used to measure the distance between simulation and the real world in a \textit{Sim-to-Real} context~\cite{zhao2020sim}. They could also be used to understand the similarities within a set of tasks in \textit{Multi-task Learning}~\cite{shuiARWG19}, or \textit{Automated Planning}~\cite{fernandez2011knowledge}.

The remainder of the paper is organized as follows. Section~\ref{sec:background} provides some preliminary concepts required to better understand the rest of the paper. Section~\ref{sec:overview} presents our proposed taxonomy, explaining how we group and categorize the different metrics. The model-based metrics are examined in Section~\ref{sec:modelbasedmetrics}, whilst the performance-based metrics are considered in Section~\ref{sec:performancebased}. In Section~\ref{sec:discussion} we discuss the surveyed methods and in Section~\ref{sec:futuredir} we identify open areas of research for future work. Finally, we conclude with Section~\ref{sec:conclusions}.

\section{Background}
\label{sec:background}

This section introduces key concepts required to better understand the rest of the paper. First, it is introduced some background in RL (Section~\ref{subsec:backrl}), then the main concepts of transfer RL are visited (Section~\ref{subsec:backtransferrl}), and finally the concepts of similarity and distance (Section~\ref{subsec:simdist}).

\subsection{Reinforcement Learning}
\label{subsec:backrl}

Typically, RL tasks are described as Markov Decision Processes (MDPs) represented by tuples in the form $\mathcal M=\langle S, A, T, R \rangle$, where $S$ is the state space, $A$  is the action space, $T : S \times A \rightarrow S$ is the transition function between states, and $R : S \times A \rightarrow \mathbb{R}$ is the reward function~\cite{sutton2011}. At each step, the agent is able to observe the current state, and choose an action according to its policy $\pi : S \rightarrow A$. The goal of the RL agent is to learn an optimal policy $\pi^*$ that maximizes the return $J(\pi)$:
\begin{equation}
    J(\pi) = \sum_{k=0}^{K}\gamma^k r_k
\label{eq:sumrewards}
\end{equation}
\noindent
where $r_k$ is the immediate reward obtained by the agent on step $k$, and $\gamma$ is the discount factor, which determines how relevant the future is (with $0 \leq \gamma \leq 1$). The interaction between the agent and the environment tends to be broken into episodes, that end when reaching a terminal state, or when a fixed amount of time has passed. With the goal of learning the policy $\pi$, Temporal Differences methods~\cite{sutton2011} estimate the sum of rewards represented in Equation~\ref{eq:sumrewards}. The function that estimates the sum of rewards, i.e., the return for each state $s$ given the policy $\pi$ is called the value-function $V^{\pi}(s) = E[J(\pi)|s_0 = s]$.  Similarly, the action-value function $Q^{\pi}(s,a) = E[J(\pi)|s_0 = s, a_0=a]$ is the estimation of the value of performing a given action $a$ at a state $s$ being $\pi$ the policy followed. The Q-learning algorithm~\cite{watkins1989} is one of the most widely used for computing the action-value function.


\subsection{Transfer Learning for Reinforcement Learning}
\label{subsec:backtransferrl}

In the transfer learning scenario we assume there is an agent who previously has addressed a set of source tasks represented as a sequence of MDPs, $\mathcal M_{1},\dots, \mathcal M_{n}$. If these tasks are somehow ``similar'' to a new task $\mathcal M_{n+1}$, then it seems reasonable the agent uses the acquired knowledge solving $\mathcal M_{1},\dots, \mathcal M_{n}$ to solve the new task $\mathcal M_{n+1}$ faster than it would be able to from scratch. Transfer learning is the problem of how to obtain, represent and, ultimately, use the previous knowledge of an agent~\cite{torrey2010transfer,taylor2009transfer}.

However, transferring knowledge is not an easy endeavour. On the one hand, we can distinguish different transfer settings depending on whether the source and target tasks share or not the state and action spaces, the transition probabilities and the reward functions. It is common to assume that the tasks share the state space and the action set, but differing the transition probabilities and/or reward functions. However, in case the tasks do not share the state and/or the action spaces, it is required to build mapping functions, $\mathcal X_{S}(s_{t})=s_{s}$, $\mathcal X_{A}(a_{t})=a_{s}$, able to map a state $s_{t}$ or action $a_{t}$ in the target task to a state $s_{s}$ or action $a_{s}$ in the source task. Such mapping functions require not only knowing if two tasks are related, but \textit{how} they are related, which means an added difficulty. On the other hand, it is required to select what type of information is going to be transferred. Different types of information have been transferred so far ranging from instance transfer (a set of samples collected in the source task) to policy transfer (i.e., the policy $\pi$ learned in the source task). Nor is this a simple task, because depending on how much and how the source and target tasks are related, it could be transferred one type of information or another.

Finally, the most ``similar'' task among $\mathcal M_{1},\dots, \mathcal M_{n}$ to solve $\mathcal M_{n+1}$ should be selected in the hope that it produces the most positive transfer. For this purpose, similarity metrics could be used, which translate into a measurable quantity of how related two tasks are. 

\subsection{Similarity and Distance Metrics}
\label{subsec:simdist}

Similarity metrics are a very important part of transfer learning, as they provide a measure of distance between tasks. Similarity functions, or their complementary distance functions, are mathematical functions that assign a numerical value to each pair of concepts or objects in a given domain. This value measures how similar  these two concepts or objects are: if they are very similar, it is assigned a very low distance, and if they are very dissimilar, it is assigned a larger distance~\cite{ontanon2020overview}.

\section{Taxonomy of Similarity Metrics for MDPs}
\label{sec:overview}

We consider there are two tasks, $\mathcal{M}_{i}$ and $\mathcal{M}_{j}$, described formally by the tuples $\mathcal{M}_{i}=\langle S_{i},A_{i},T_{i},R_{i}\rangle$ and $\mathcal{M}_{j}=\langle S_{j},A_{j},T_{j},R_{j}\rangle$, where they could share (or not) the state space, the action space, or the transition and reward dynamics. 


\begin{definition}
Given two tasks $\mathcal{M}_{i}$ and $\mathcal{M}_{j}$, we define a task distance metric as a heuristic function $d(\mathcal{M}_{i}, \mathcal{M}_{j}) \rightarrow [0,\infty)$, such that if $d(\mathcal{M}_{i}, \mathcal{M}_{j}) < d(\mathcal{M}_{k}, \mathcal{M}_{j})$, then $\mathcal{M}_{i}$ is considered more \textnormal{similar} to $\mathcal{M}_{j}$ than $\mathcal{M}_{k}$.\footnote{It should be noted that in the context of task similarity such a measure of distance need not be a ``metric'' in the mathematical sense of this term, but it needs to allow us to define an order between tasks.}
\label{def:def1}
\end{definition}

Definition~\ref{def:def1} allows us to use the function $d(\cdot, \cdot)$ to obtain an ordering between tasks in such a way that we can select the more \textit{similar} one. Ideally, the concept of similarity should be related to the concept of positive transfer: the smaller the distance $d(\mathcal{M}_{i}, \mathcal{M}_{j})$, the greater the positive transfer. Additionally, $d(\mathcal{M}_{i}, \mathcal{M}_{j})$ should be computed \textit{before} or, at least, \textit{during} the transfer experiment, in order to select an adequate task to use in transfer. However, the literature proposes different ways to compute  $d(\mathcal{M}_{i}, \mathcal{M}_{j})$. 

\begin{figure}
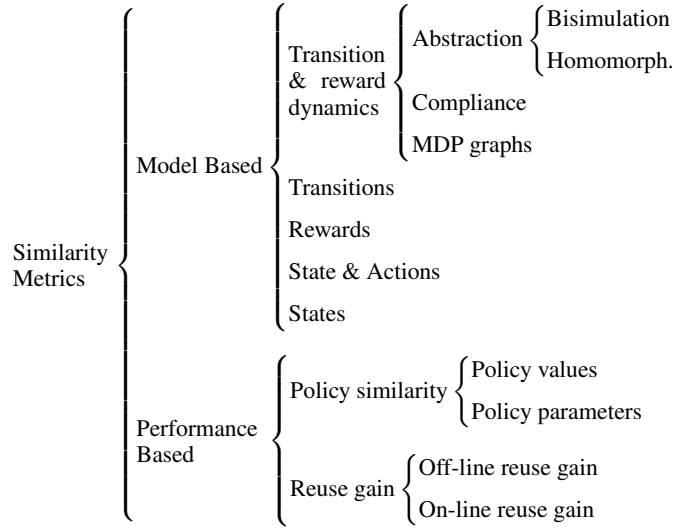

\centering
\small
    $\parbox{1.3cm}{Similarity Metrics} 
    \begin{cases}
        \text{Model Based} 
        \begin{cases}
        \parbox{1.3cm}{Transition \& reward dynamics}
        \begin{cases}
            \text{Abstraction}
            \begin{cases}
            \text{Bisimulation} \\
            \noalign{\vskip4pt}
            \text{Homomorph.}\\
            \end{cases}\\
            \noalign{\vskip4pt}
            \text{Compliance}\\
            \noalign{\vskip4pt}
            \text{MDP graphs}\\
        \end{cases}\\
            \noalign{\vskip4pt}
            \text{Transitions}\\
            \noalign{\vskip4pt}
            \text{Rewards}\\
            \noalign{\vskip4pt}
            \text{State \& Actions}\\
            \noalign{\vskip4pt}
            \text{States} \\
        \end{cases} \\
        \noalign{\vskip9pt}
        \parbox{1.7cm}{Performance Based} 
        \begin{cases}
            \text{Policy similarity}
            \begin{cases}
                \text{Policy values}\\
                \noalign{\vskip4pt}
                \text{Policy parameters}\\
            \end{cases}\\
            \noalign{\vskip9pt}
            \text{Reuse gain}
            \begin{cases}
                \text{Off-line reuse gain} \\
                \noalign{\vskip4pt}
                \text{On-line reuse gain}\\
            \end{cases}
        \end{cases}
    \end{cases}$
    \caption{Overview of the similarity metrics considered in this survey}
    \label{tab:metricsTable}
\end{figure}

In this paper, we consider two main trends for the computation of the distance metric $d(\mathcal{M}_{i}, \mathcal{M}_{j})$. Such trends are depicted in Figure~\ref{tab:metricsTable}. The first one measures the structural or model similarities between the given MDPs. The second measures the similarities by using the performance of the learning agent in both the source and target tasks.

\section{Model-based Metrics}
\label{sec:modelbasedmetrics}

As regards the first, model-based metrics measure the degree of similarity between a source and a target task by using their corresponding MDP models (i.e., states, actions, transition and rewards dynamics). There are several alternatives to this model-based metrics depending on what components of the MDPs are taken into account. In our proposed taxonomy, we categorize these metrics in five groups: (i) transition and reward dynamics, (ii) transitions, (iii) rewards, (iv) state and actions and (v) states.

\subsection{Transition \& reward dynamics}

They require complete knowledge of the MDP models both of the source and the target tasks. We distinguish three ways of computing similarity metrics using such a complete knowledge: (i) by a sort of metrics based on state abstraction (or state aggregation) techniques~\cite{li2006towards,ferns2004metrics,ferns2012metrics,castro2020}, (ii) by compliance metrics~\cite{lazaricRB08,phdthesislazaric,fachantidis2015transfer,phdthesisfachantidis} and (iii) by metrics based on the construction of \textit{MDP graphs}~\cite{kuhlmann:ecml07,ijcai2019-511}.

\subsubsection{State abstraction}

In RL, a common practice is to aggregate states in order to obtain an \textit{abstract} description of the problem, i.e., a more compact and easier representation of the task to work with~\cite{li2006towards}. These approaches are based on the same common principle: if a number of states are considered to be \textit{similar}, they can be aggregated as a single one. This same principle can also be used to compute the \textit{similarity} between two states belonging to different MDPs. In this paper, we survey two of these methods which actually have been used for transfer in RL: bisimulation~\cite{ferns2004metrics,ferns2012metrics,castro2011automatic,song16}, and homomorphism~\cite{ravindran2002model,Sorg2009}.

\begin{definition}
    Given two MDPs $\mathcal M_{i}$ and $\mathcal M_{j}$, we define $d(\mathcal M_{i}, \mathcal M_{j}) = d(S_{i}, S_{j})$, where $d(S_{i}, S_{j})$ measures the distance between $S_{i}$ and $S_{j}$ by computing the bisimulation or homomorphism distances of the individual state pairs, $d(s_{i},s_{j})$, $\forall s_{i},s_{j} \in S_{i} \times S_{j}$.
\end{definition}

Bisimulation metrics compute $d(s_{i},s_{j})$ by comparing the transition and reward dynamics of $s_{i}$ and $s_{j}$: the more similar the reward and transition structures of $s_{i}$ and $s_{j}$ are, the smaller $d(s_{i},s_{j})$. One of the shortcomings of the bisimulation metrics is that they require both MDPs, $\mathcal M_{i}$ and $\mathcal M_{j}$, to have the same action sets, $A_i = A_j$. However, such a shortcoming has been successfully addressed by the homomorphism metrics which are able to deal with different action spaces, $A_i \neq A_j$~\cite{ravindran2002model,Sorg2009}. Once the distance between all state pairs in $\mathcal M_{i}$ and $\mathcal M_{j}$ is computed, it is required to composite them to compute the distance $d(\mathcal M_{i}, \mathcal M_{j})$ between the two MDPs. It could not be appropriate to simply accumulate or average the distances between all different state pairs, so it is typical to define $d(S_{i},S_{j})$ as a function which measure the distance between the sets corresponding to the state spaces $S_{i}$ and $S_{j}$, e.g., the Hausdorff, or the Kantorovich function~\cite{song16}, although other metrics between sets would also be possible~\cite{conci2018distance}.

\subsubsection{Compliance}
    
The \textit{compliance} measure is defined as the probability of a sample $\langle s,a,s',r\rangle$ in a target task $\mathcal M_{j}$ of being generated in the source task $\mathcal M_{i}$~\cite{lazaricRB08,phdthesisfachantidis,fachantidis2015transfer}. Therefore, it is easy to deduce that the compliance between the entire target task and the entire source task allows us to measure the similarity between the two tasks.
\begin{definition}
    Given two MDPs, $\mathcal M_{i}$ and $\mathcal M_{j}$, and two sets of experience tuples generated $D_{\mathcal M_{i}}$ and $D_{\mathcal M_{j}}$ gathered from $\mathcal M_i$ and $\mathcal M_j$, the compliance between $\mathcal M_{i}$ and $\mathcal M_{j}$ is computed as:
    \begin{equation}
         \Lambda = \frac{1}{n}\sum_{t=0}^{n} P(\sigma_{t}|D_{\mathcal M_i})
    \label{eq:complianceent}
    \end{equation}
    \noindent where $n$ is the number of samples in $D_{\mathcal M_{j}}$, and $\sigma_{t}$ is the $t$-th tuple in $D_{\mathcal M_{j}}$.
\label{def:complianceent}
\end{definition}

$\Lambda$ is not strictly a distance metric but a probability: the more likely the samples of the target task are generated in the source task, the closer $\Lambda$ to $1$. Therefore, \textit{compliance} could be used to obtain a distance metric between MDPs like the ones this survey is looking for, e.g., $d(\mathcal M_{i}, \mathcal M_{j}) = 1 - \Lambda$.
    
    
\subsubsection{MDP graphs}
    
They are based on the construction of graphs that represent the transition and the reward functions both of the source task and the target task. Then, they find structural similarities between tasks based on graph-similarity or graph-matching algorithms. 

\begin{definition}
    Given two MDPs $\mathcal M_{i}$ and $\mathcal M_{j}$ and their corresponding alternative representation as graphs, $G_{\mathcal M_{i}}$ and $G_{\mathcal M_{j}}$, we define $d(\mathcal M_{i}, \mathcal M_{j})$ as inversely related to $\Phi(G_{\mathcal M_{i}}, G_{\mathcal M_{j}})$, where $\Phi(\cdot,\cdot)\rightarrow[0,\infty)$ is a function that measures the structural similarity between $G_{\mathcal M_{i}}$ and $G_{\mathcal M_{j}}$.
\end{definition}
    
\citeauthor{ijcai2019-511} (\citeyear{ijcai2019-511}) represents the MDPs as bipartite directed graphs, which permits the computation of structural similarity measures between them. Although several of these structural functions $\Phi(\cdot,\cdot)$ could be used such as RoleSim~\cite{jin2014scalable} or MatchSim~\cite{lin2012matchsim}, Wang et al. are based on SimRank~\cite{SimRank}, which basic idea is that \textit{two nodes are similar iff their neighbors are similar}.
On the other hand, \citeauthor{kuhlmann:ecml07} (\citeyear{kuhlmann:ecml07}) represents the MDPs as rule graphs instead of bipartite graphs for the particular problem of General Game Playing~\cite{genesereth2005general}. Such a rule graph is an accurate abstraction of the MDP problem, and can be properly compared to other rule graphs using an isomorphic function as $\Phi(\cdot,\cdot)$~\cite{mckay2014practical}. Finally, \citeauthor{AAAI06-yaxin} (\citeyear{AAAI06-yaxin}) models the problem as a Qualitative Dynamic Bayes Network (QDBN). This assigns types to the nodes and edges, providing additional characteristics to compare.
    
        
\subsection{Transitions}

The metrics considered in this category use tuples in the form $\langle s,a,s'\rangle$ to measure the similarity between MDPs. 
\begin{definition}
    Given two tasks $\mathcal M_{i}$ and $\mathcal M_{j}$, and two sets $\mathcal D_{\mathcal M_{i}}$ and $\mathcal D_{\mathcal M_{j}}$ of experience tuples in the form $\tau = \langle s,a,s'\rangle$ gathered from $\mathcal M_{i}$ and $\mathcal M_{j}$, we can define the distance between them as $d(\mathcal M_{i}, \mathcal M_{j})=d(\mathcal D_{\mathcal M_{i}}, \mathcal D_{\mathcal M_{j}})$.
\end{definition}

\citeauthor{ammar14} (\citeyear{ammar14}) use $\mathcal D_{\mathcal M_{j}}$ to build an RBM model which describes the transitions in $\mathcal M_{j}$ in a richer feature space. Then, they feed the tuples $\tau_{k} \in \mathcal D_{\mathcal M_{i}}$ into this RBM model to obtain a reconstruction $\tau'_{k}$. Afterward, they compute the Euclidean distance between $\tau_{k}$ and $\tau'_{k}$.  Similarly, \citeauthor{taylor2008autonomous} (\citeyear{taylor2008autonomous}) use $\mathcal D_{\mathcal M_{j}}$ to learn a one-step transition model $M(s,a) \rightarrow s'$. Afterward, for each tuple $\tau_{k}=\langle s,a,s'\rangle$ in $\mathcal D_{\mathcal M_{i}}$, they compute the Euclidean distance between $s'$ and $M(s,a)$. In both cases, the distance $d(\mathcal D_{\mathcal M_{i}}, \mathcal D_{\mathcal M_{j}})$ is computed as the average of the Euclidean distances obtained for each $\tau_{k} \in D_{\mathcal M_{i}}$.


\subsection{Rewards}

The metrics can also measure the similarity between MDPs according to the distance between their reward dynamics.

\begin{definition}
    Given two MDPs $\mathcal M_{i}$ and $\mathcal M_{j}$, we define the distance between them as $d(\mathcal M_{i}, \mathcal M_{j})=d(R_{i}, R_{j})$.
\end{definition}

For instance, \citeauthor{1555955} (\citeyear{1555955}) computes $d(R_{i}, R_{j})$ as  $d(R_{i}, R_{j})=\frac{1}{n} \sum_{s \in S} \sum_{a \in A} (R_{i}(s,a) - R_{j}(s,a))^2$, where $n$ is the total number of state-action pairs in the source and the target task.
Instead, \citeauthor{tao2021repaint} (\citeyear{tao2021repaint}) assumes the reward functions are a linear combination of some common \textit{features} $\phi(\cdot,\cdot)$, $R_{i}(s,a)=\phi(s,a)^T w_{i}$ and $R_{j}(s,a)=\phi(s,a)^{T} w_{j}$, and then  use the cosine distance function between $w_{i}$ and $w_{j}$ to compute $d(R_{i}, R_{j})$.
Finally, in the literature there are several ways of computing $d(R_{i}, R_{j})$~\cite{gleave2020quantifying}.


\subsection{State \& Action Spaces}

In this case, the distance between MDPs is computed as the distance between the state-action spaces. 

\begin{definition}
    Given two MDPs $\mathcal M_{i}$ and $\mathcal M_{j}$, and their corresponding state-action spaces, $S_{i} \times A_{i}$ and $S_{j} \times A_{j}$, we define the distance as $d(\mathcal M_{i}, \mathcal M_{j})=d(S_{i} \times A_{i}, S_{j} \times A_{j})$.
\end{definition}

For instance, \citeauthor{narayan2019effects} (\citeyear{narayan2019effects}) compute this distance as the averaged difference between the corresponding state-action transition distributions of the two tasks.
Instead, \citeauthor{10.1007/978-3-540-87481-2_32} (\citeyear{10.1007/978-3-540-87481-2_32}) compute the Euclidean distance between state-action pairs in the source and target tasks. This work is not focused on the construction of a distance measure between MDPs, but such Euclidean distances between all the state-action pairs could be composited to obtain a sort of similarity metric.

\subsection{States}

Finally, the metrics in this category use the state space in both the source and target tasks to compute the similarity between them. 

\begin{definition}
    Given two MDPs $\mathcal M_{i}$ and $\mathcal M_{j}$, we can define the distance between them as $d(\mathcal M_{i}, \mathcal M_{j}) = d(S_{i}, S_{j})$, where $d(S_{i}, S_{j})$ measures the distance between $S_{i}$ and $S_{j}$.
\end{definition}

\citeauthor{svetlik17} (\citeyear{svetlik17}) propose to compute $d(S_{i}, S_{j})$ as the relation between the applicability of the value function (measured as the number of states in $\mathcal M_{i}$ that also appears in $\mathcal M_{j}$), and the experience required to learn in $\mathcal M_{j}$ (measured as the difference of size between $S_{j}$ and $S_{i}$). Another area of research that is relevant in this category is that of case based reasoning (CBR)~\cite{aamodt1994case}. RL approaches based on CBR use a similarity function between the states in the target task and the states stored in a case base corresponding to a previous source task~\cite{celiberto2011using}. Such similarity function could be used to measure the similarity between the state spaces, hence, the similarity between the two tasks.

\section{Performance-based Metrics}
\label{sec:performancebased}

As regards the second major category in the proposed taxonomy, they are based on the \textit{performance} of the agents in the source task and the target task, where this \textit{performance} can be related to the policies themselves learned by the agents in these tasks, or to the reuse gain an agent obtains reusing the knowledge of a source task in a target task. So, we distinguish two different approaches to overcoming the problem of computing such a performance-based similarities: (i) by the policy similarity and (ii) by the reuse gain obtained transferring the knowledge from a source task to the target task.

\subsection{Policy similarity}

They are based on the use of the learned value function $V^{\pi}$ or the action-value function $Q^{\pi}$, or equivalently, on the behavioral policies $\pi$ obtained in the source task and the target task. Therefore, these metrics require the full (or partial) learning of these policies before the computation of the similarity between tasks. Such a comparison can be conducted in two different ways depending on what is being compared: (i) the policy values, or (ii) the policy parameters.

\subsubsection{Policy values}

In this case, the comparison is conducted by observing the specific $q$-values or $v$-values of the $Q^{\pi}$-function or the $V^{\pi}$-function of the source and target tasks. Therefore, in this case, it is really being measured the degree of similarity of the policies obtained in both tasks.

\begin{definition}
Given two tasks $\mathcal M_{i}$ and $\mathcal M_{j}$ and the $q$-values of $Q^{\pi_{i}}$ and $Q^{\pi_{j}}$ learned in these tasks, we define
$d(\mathcal M_{i}, \mathcal M_{j}) = d(Q^{\pi_{i}},Q^{\pi_{j}})$. This metric is transformed into $d(\mathcal M_{i}, \mathcal M_{j}) = d(V^{\pi_{i}},V^{\pi_{j}})$ if $V^{\pi_{i}}$ and $V^{\pi_{j}}$ are computed instead.
\end{definition}

\citeauthor{1555955} (\citeyear{1555955}) propose to compute  $d(V^{\pi_{i}},V^{\pi_{j}})$ as the number of states with identical maximum $v$-value, and $d(Q^{\pi_{i}},Q^{\pi_{j}})$ as the mean squared error of the $q$-values in $Q^{\pi_{i}}$ and $Q^{\pi_{j}}$. Instead, \citeauthor{Zhou2020} (\citeyear{Zhou2020}) compute $d(Q^{\pi_{i}},Q^{\pi_{j}})$ deriving latent structures of tasks and ﬁnding matches between $Q^{\pi_{i}}$ and $Q^{\pi_{j}}$. 
    
\subsubsection{Policy Approximation Parameters}

In RL, the value function $V^{\pi}$ or the action-value function $Q^{\pi}$ usually are represented or approximated with a parameter vector $\theta$. Intuitively, the metrics within this category compare the particular weights of the parameter vectors corresponding to the value functions of the source task and the target task in order to measure the similarity between them. 

\begin{definition}
Given two tasks $\mathcal M_{i}$ and $\mathcal M_{j}$ and the parameterize functions $Q^{\theta_{i}} \approx Q^{\pi_{i}}$ and $Q^{\theta_{j}} \approx Q^{\pi_{j}}$ learned in these tasks, we consider $d(\mathcal M_{i}, \mathcal M_{j}) = d(\theta_{i},\theta_{j})$.
\end{definition}

The distance $d(\theta_{i},\theta_{j})$ can be computed using the cosine similarity between two non-zero vectors~\cite{Karimpanal2018}. 
Such an approach opens the door to the comparison of other policy representations~\cite{ferrante2008transfer}.

\subsection{Reuse gain}

In these techniques, the level of similarity is an approximation to the \textit{advantage} gained by using the knowledge in one source task to speed up the learning of another target task~\cite{1555955,carroll2005task,taylor2009transfer}. 

\begin{definition}
    Given two tasks $\mathcal M_{i}$ and $\mathcal M_{j}$, and $g(\mathcal M_{i}, \mathcal M_{j})$ which denotes the gain of transferring the knowledge learned in $\mathcal M_{i}$ to $\mathcal M_{j}$, we can consider $d(\mathcal M_{i}, \mathcal M_{j})$ as inversely related to $g(\mathcal M_{i}, \mathcal M_{j})$.
\end{definition}

Therefore, it is important to bear in mind that these metrics actually require that the transfer experiment be entirely or partially run before measuring the degree of similarity between tasks. However, regardless of the particular technique used to compute such a reuse gain, the higher the reuse gain, the greater the similarity between the tasks. In this paper, we distinguish two approaches within this category depending on whether the reuse gain is computed \textit{after} or \textit{during} the transfer process: (i) off-line reuse gain, and (ii) on-line reuse gain.

\subsubsection{Off-line reuse gain}
    
In this case, the reuse gain $g(\mathcal M_{i}, \mathcal M_{j})$ is estimated as the difference in performance between the learning process with and without transfer, and once the learning processes are considered to be finished~\cite{carroll2005task,mahmud13,sinapov15,zhan2016theoreticallygrounded}. Such gain $g(\mathcal M_{i}, \mathcal M_{j})$ can be computed as the \textit{jumpstart}~\cite{sinapov15,carroll2005task}, the time to convergence~\cite{carroll2005task}, the asymptotic performance~\cite{mahmud13}, although other metrics such as the total reward or the transfer ratio could be also used~\cite{taylor2009transfer}.
    
        
\subsubsection{On-line reuse gain}

On the contrary, in these approaches, the reuse gain is estimated \textit{on-line} at the same time that the policy in the target task is computed~\cite{fernandez12,azar2013regret,li2017optimal}. It is important to bear in mind that the \textit{on-line} computation of this gain only makes sense if during the learning process we have several transfer sources to choose from. At the beginning of the learning process, these approaches have at its disposal the knowledge learned in solving a set of previous tasks $\{\mathcal M_{1}, \dots, \mathcal M_{n}\}$ to learn the new task $\mathcal M_{j}$. During learning, they compute $g(\mathcal M_{i}, \mathcal M_{j})$ of each past task $\mathcal M_{i} \in \{\mathcal M_{1}, \dots, \mathcal M_{n}\}$. To do that, they transfer the knowledge acquired solving $\mathcal M_{i}$ to $\mathcal M_{j}$ during a limited number of episodes $m$. Then, $g(\mathcal M_{i}, \mathcal M_{j})$ is computed as the average reward obtained during those $m$ episodes~\cite{fernandez12,azar2013regret}. Once all gains are computed, it is possible to decide \textit{on-line} which is the closest task to $\mathcal M_{j}$ within $\{\mathcal M_{1}, \dots, \mathcal M_{n}\}$, so that the knowledge of the selected closest task can have a greater influence on learning about the policy in $\mathcal M_{j}$. 
    

\section{Discussion}
\label{sec:discussion}

From a transfer point of view, the ultimate goal of all similarity metrics is in some way to \textit{predict} the relative advantage that would be gained by using a source task in a target task. The more similar the source and target tasks are, the greater the \textit{positive transfer}. The correct selection of a distance metric should be carried out attending to four dimensions: nature of the state-action space, amount of information available on the tasks, computation moment, and transfer technique.

Obviously, the selection of the metric depends on the nature of the state-action space. Some approaches typically require enumerating all the states both in the source and the target tasks~\cite{castro2020,Zhou2020}, but such full state enumeration is impractical for large state spaces, and impossible for continuous state spaces. Other methods require that both tasks have the same state-action space~\cite{lazaricRB08,azar2013regret}, which is not true in most of the cases. However, the latter can be partially alleviated by the construction of inter-task mapping functions, $\mathcal X_{S}$ and $\mathcal X_{A}$.

Another issue that needs to be addressed is how much information we have about the tasks to solve. Most model-based approaches require \textit{prior} full information (or accurate approximations) about the transition and reward dynamics~\cite{castro2010using,lazaricRB08,wang2019research}, or about the size of the state and action spaces ~\cite{svetlik17,carroll2005task}. In this sense, performance-based metrics have a clear advantage over model-based ones: in general, performance-based metrics require less \textit{a priori} information about the task to be solved, although as a counterpoint they need to fully or partially run the transfer experiment to obtain the distance measurement. 

This leads us to the third issue: the computation moment. Ideally, the computation of the similarity metric should be before or, at least, during the transfer. Off-line reuse gain approaches are undoubtedly the best method for measuring similarity between two tasks: they produce such a measure \textit{after} the transfer 
experiment has been run, in such a way that we can compute the real gain. However, if the point is to use the task similarity measure to choose a task to use in transfer, these metrics are useless. In this case, model-based metrics take advantage over performance-based metrics: they allow to compute the metrics \textit{before} the transfer process. These metrics can be used to choose the most similar MDP before transfer, but as far as we know there is no theoretical guarantees that the most similar MDP is \textit{similar} enough to produce a positive transfer. By contrast, the metrics based on the on-line reuse gain are at the point half-way between both. They allow to compute the similarity metric \textit{during} transfer, so that depending on the similarity of the source task, it will introduce a greater or lesser \textit{exploration bias} in the learning process of the new task. 

Finally, the success of a metric is intimately tied with the transfer technique that uses it~\cite{carroll2005task,1555955}. This means that there is probably no one best universal metric that works with all transfer techniques and problems, in the same way that there is no one best universal transfer technique. Since each metric can capture different types of similarity and each transfer technique induces different bias in the learning process, the question of selecting the best metric turns into finding the correct metric for a transfer technique to be applied to a particular task.

\section{Future Directions}
\label{sec:futuredir}

The previous discussion points out several future directions. On the one hand, since there is not a best metric, it would be useful to use several of them. For instance, model-based metrics can be used to return a useful approximation of task similarity before the tasks are learned, although this measure can be adapted \textit{on-line} during the learning process so that the bias that the source task induces in the exploration process is adjusted dynamically. On the other hand, given that different metrics compute different types of similarity (i.e., model-based metrics measure \textit{structural} similarities, whilst performance-based metrics measure \textit{performance} similarities), the agent can be equipped with the ability not only to determine which source task to use, but also which transfer technique to use given the type of similarity between the source and target tasks.

Another interesting line of research is that based on building semantic representations of the tasks through domain-dependent features. For instance, we can define a particular Pac-Man task from features like the number of ghosts, behavior of the ghosts, or the type of the maze, and use these features to build a similarity metric between different Pac-Man tasks. In fact, one may heuristically combine structural, performance, but also semantic similarity aspects into a same metric. Thus, it could be obtained a metric more aligned with the way in which humans decides what is similar, since humans analyze the similarity between concepts or objects from different perspectives~\cite{kemmerer2017}.

Finally, transferring learned models from simulation to the real world remains one of the hardest problems in control theory~\cite{zhao2020sim}. In this case, similarity metrics can help to answer how similar simulations and the actual world are. They could be used to provide theoretical guaranties that ensure the learned policies transferred from simulation to the actual world will perform as required, or to deﬁne mechanisms to tune/modify the simulated environments, so the gap between the simulated world and the actual one decreases.


\section{Concluding Remarks}
\label{sec:conclusions}

This paper contributes a compact and useful taxonomy of similarity metrics for Markov Decision Processes. The leaves of the taxonomy have been used to provide a literature review that surveys the existing work. We differentiated between model-based and performance-based metrics, depending on whether a \textit{structural} or \textit{performance} criterion has been used in its creation. The proposed taxonomy permits to organize clearly the different similarity metrics, or find commonalities between them. This can help the reader to choose similarity metrics for their tasks, or even define their own. We also discussed different selection criteria and some promising future research directions.


\section*{Acknowledgements}

This research was funded in part by JPMorgan Chase \& Co. Any views or opinions expressed herein are solely those of the authors listed, and may differ from the views and opinions expressed by JPMorgan Chase \& Co. or its affiliates. This material is not a product of the Research Department of J.P. Morgan Securities LLC. This material should not be construed as an individual recommendation for any particular client and is not intended as a recommendation of particular securities, financial instruments or strategies for a particular client. This material does not constitute a solicitation or offer in any jurisdiction.\\

\noindent This work has been supported by the Madrid Government (Comunidad de Madrid-Spain) under the Multiannual Agreement with UC3M in the line of Excellence of University Professors (EPUC3M17), and in the context of the V PRICIT (Regional Programme of Research and Technological Innovation).

\bibliographystyle{named}
\bibliography{ijcai21}

\end{document}